\documentclass[letterpaper, 10 pt, conference]{ieeeconf}

\usepackage{geometry}
\IEEEoverridecommandlockouts 

\usepackage{ifpdf}

\usepackage{dblfloatfix} 

\usepackage{cite}

\usepackage[pdftex]{graphicx}
\usepackage{multirow}
\usepackage{booktabs}

\usepackage{amsmath}
\usepackage{amsfonts}
\usepackage{amssymb}

\DeclareMathOperator*{\argmax}{arg\,max}

\DeclareMathOperator{\varop}{Var}
\DeclareMathOperator{\mse}{MSE}
\DeclareMathOperator{\me}{ME}

\newcommand{\mean}[1]{\mathbb{E}\left[#1\right]}
\newcommand{\var}[1]{\varop \left[#1\right]}

\usepackage[english]{babel}

\newtheorem{theorem}{Theorem}
\newtheorem{lemma}[theorem]{Lemma}

\newtheorem{example}{Example}

\usepackage[caption=false,font=footnotesize]{subfig}

\makeatletter
\let\NAT@parse\undefined
\makeatother
\usepackage[colorlinks=true, allcolors=blue]{hyperref}

\let\labelindent\relax
\usepackage{enumitem}

\newcommand{\ourplain}{\textsc{MexGen}}
\newcommand{\our}{{\small \ourplain}}

\newcommand{\tableref}[1]{Table~\ref{#1}}
\newcommand{\figref}[1]{Figure~\ref{#1}}

\newcommand{\secref}[1]{$\S$~\ref{#1}}
\newcommand{\speedup}[0]{\(7.6\times\)}

\begin{document}

\title{\LARGE \bf \ourplain: An Effective and Efficient Information Gain Approximation for Information Gathering Path Planning}

\author{Joshua Chesser$^{1}$, Thuraiappah Sathyan$^{2}$  and Damith C. Ranasinghe$^{1}$
\thanks{*This work was supported by grant funding from Lockeed Martin, Australia, Australian Research Council (ARC) grant LP200301507 and the Australian Government’s Research Training Program Scholarship (RTPS).}
\thanks{$^{1}$Joshua Chesser and Damith C. Ranasinghe are with the School of Computer \& Mathematical Sciences, The University of Adelaide, SA 5005, Australia.
        {\tt\small joshua.chesser@adelaide.edu.au, damith.ranasinghe@adelaide.edu.au}}%
\thanks{$^{2}$Sathyan Thuraiappah is with STELaRLab, Lockheed Martin Australia.
        {\tt\small sathyan.thuraiappah@global.lmco.com}}%
}

\maketitle
\thispagestyle{empty}
\pagestyle{empty}

\begin{abstract}
    Autonomous robots for gathering information on objects of interest has numerous real-world applications because of they improve efficiency, performance and safety. Realizing autonomy demands online planning algorithms to solve sequential decision making problems under \textit{uncertainty}; because, objects of interest are often dynamic, object state, such as location is not directly observable and are obtained from noisy measurements. Such planning problems are notoriously difficult due to the combinatorial nature of predicting the \textit{future} to make optimal decisions. For information theoretic planning algorithms, we develop a \textit{computationally efficient} and \textit{effective} approximation for the difficult problem of predicting the \textit{likely sensor measurements from uncertain belief states}. The approach more accurately predicts information gain from information gathering actions. Our theoretical analysis \textit{proves} the proposed formulation  achieves a lower prediction error than the current efficient-method. We demonstrate improved performance gains in radio-source tracking and localization problems using extensive simulated and field experiments with a multirotor aerial robot. 
\end{abstract}

\section{Introduction}
Autonomous sensing can robotize dull, dangerous, labour-intensive information gathering tasks,  
such as search and rescue~\cite{schedl_autonomous_2021},  exploration~\cite{tao_seer_2023} and wildlife tracking for conservation~\cite{vander2014,vander2016,cliff_robotic_2018,fei_conservationbots_2023,bayram_gathering_2016, yilmaz_particle_2023}. Typical systems are built with agents capable of performing various \textit{information gathering actions}, such as changing sensor positions 
to optimize the \textit{sensor-to-target geometry}
to \textit{gain information}, e.g. position and velocity, on objects of interest. Building autonomous 
agents for the task requires solving a planning problem---a sequential decision making process where each agent repeatedly selects the next best information gathering action.
Critically, to perform a task effectively, planning methods must deal with uncertainty---future actions need to: i)~rely on \textit{imperfect} state information, such as positions of wildlife, gathered by \textit{noisy} measurements; and ii)~account for non-deterministic system dynamics, such as movements and behaviors of tracked wildlife.

Partially observable Markov decision process (POMDP) formulations are widely used for \textit{online} planning problems under \textit{uncertainty}~\cite{cliff_robotic_2018,nguyen_trackerbots_2019,dressel_hunting_2019,araya_pomdp_2010,sunberg_online_2018,fischer_information_2020,sztyglic_speeding_2022,ott_sequential_2023, burks_optimal_2019, cai_hyp-despot_2021,Nguyen_Rezatofighi_Vo_Ranasinghe_2020}. Our work focuses on information theoretic POMDP planning problems for object tracking. The goal is to select the optimal  actions to obtain the most informative measurements to accurately estimate the state of dynamic objects of interest.

Optimal planning requires tackling the computationally challenging problem of estimating the information gain of future actions with respect to the large space of possible future measurements. An agent must consider the result of \textit{every possible measurement} that could be obtained after every possible action at each decision step over a look-ahead horizon. The result is an exponentially increasing number of possible action-measurement sequences as the horizon increases; notably, solving POMDPs optimally is PSPACE-complete~\cite{papadimitriou_complexity_1987}. Hence, a significant challenge is formulating suitable approximations for the problem to achieve computationally efficient, yet effective planning decisions for information gathering tasks. 

\begin{figure}[!t]
    \centering
    \subfloat{\includegraphics[width=\linewidth, trim={0 0 0 6mm},clip]{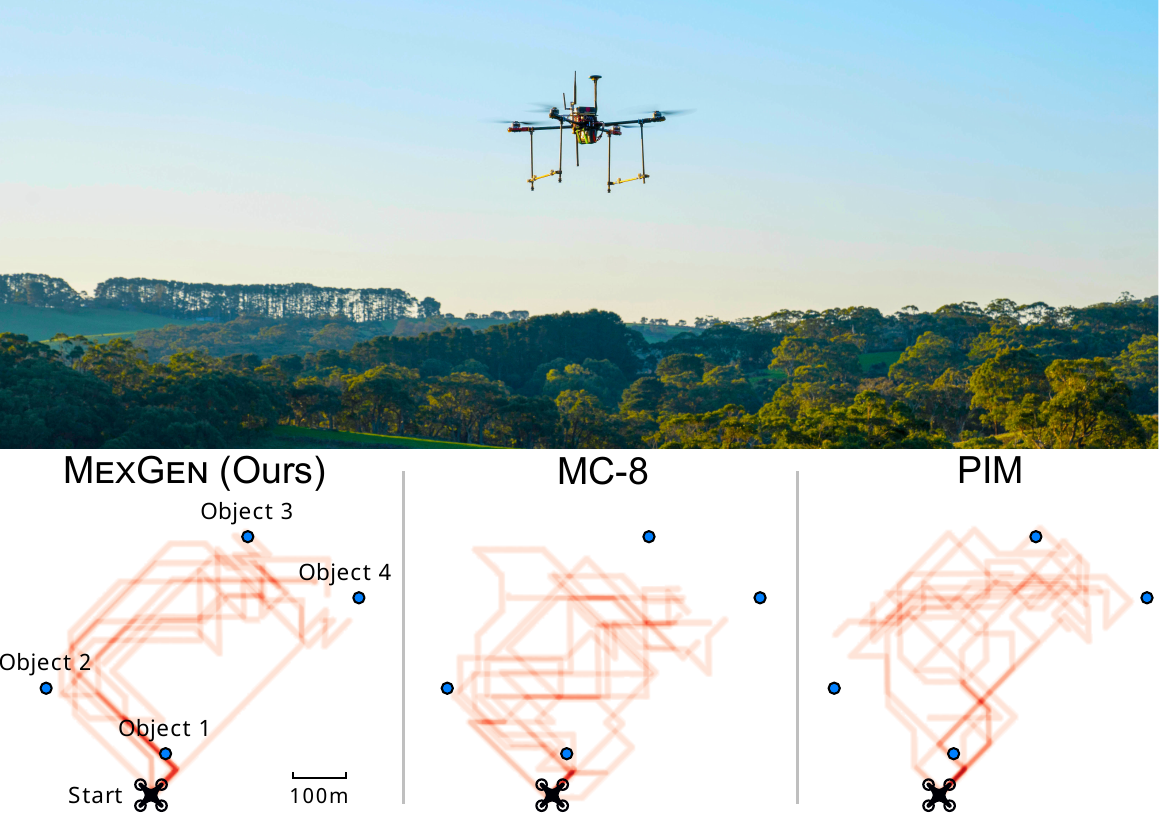}\label{fig:field_teaser}}
    \caption{Trajectory heat maps from field experiments. We compared the current efficient method (PIM) and a computationally expensive method (MC-8) for computing information gain from possible future measurements with ours for a radio source localization task. Trajectory planning decisions, i.e. information gathering actions, are more effective with ours---resulting in a \(40\)\% reduction in mission completion times in the best case (vs. PIM, the next best method) and visibly \textit{shorter} paths leading to consistently localizing the closest object and navigating to reduce the uncertainty of the next object (see \secref{sec:field}). The  demonstration video is at: \url{https://youtu.be/XrsCC6MkaB4}}
    \label{fig:field_results}
    \vspace{-3mm}
\end{figure}
Existing work approximate the information gain with: i)~a subset of possible measurements sampled for each action from the future belief state with Monte Carlo (MC) methods~\cite{beard_void_2017, dressel_hunting_2019, sunberg_online_2018, nguyen_trackerbots_2019, fischer_information_2020, sztyglic_speeding_2022}. The method's effectiveness depends on the number of samples but increasing samples increases computational costs exponentially; or ii)~a single noiseless measurement generated from a likely future state approximation---the predicted ideal measurement (PIM)~\cite{ristic_note_2011, cliff_robotic_2018, yilmaz_particle_2023}. The methods assume the true future state is a mode of the predicted belief state distribution and approximates it with a point estimate. 

\begin{figure*}[b]
    \centering
    \includegraphics[width=1.0\linewidth]{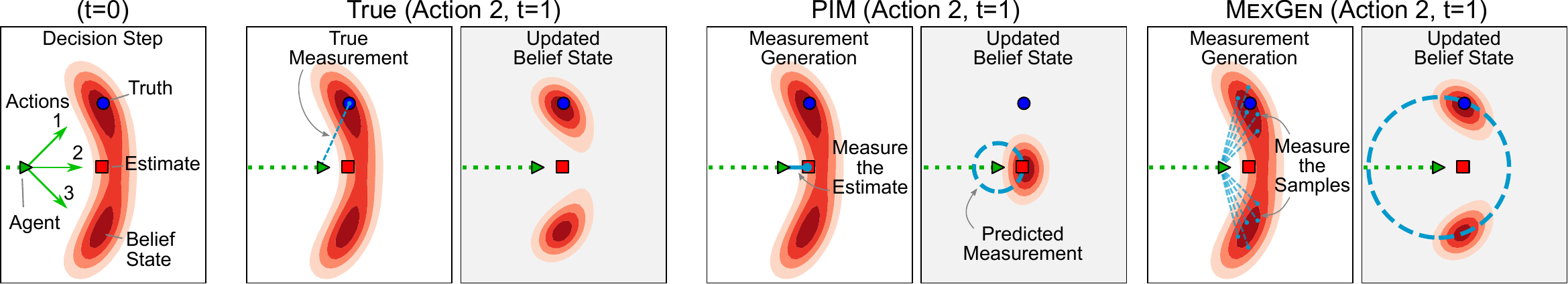}
   
    \caption{Consider an agent with an onboard sensor for acquiring noisy range measurements tasked with tracking the position an object of interest (e.g. radio source). At time $t=0$, the agent evaluates \textit{Action~2}'s reward out of three possible actions over a look ahead horizon of $t=1$. The red contours depict the belief state probability density. In \textit{True}, we update the belief state with a  measurement from the ground truth object position at $t=1$ for comparison. \textit{PIM} and \ourplain\ tiles show the \textit{Updated Belief State} \textit{given} the predicted \textit{range} measurement generated at \(t=1\). Now, PIM, the existing computationally efficient approximation, incorrectly predicts an unrealistic measurement from an estimated future estate of an object. The result is  an over confident future object belief state that is significantly different from the \textit{Truth}.  In contrast, \ourplain\ uses the expected measurement and updates the belief state to predict the future state; this result compares well with the \textit{True} state at $t=1$ in contrast to PIM and leads to a better approximation of information gain from \textit{Action~2}.}
    \label{fig:pims_example}
\end{figure*}

This study formulates a computationally efficient, yet more effective approximation for information gain. We propose computing information gain of each information gathering action at each decision step using the \textit{expected} measurement generated from the predicted belief state to yield the updated belief state---or future belief state. Effectively, we exploit the asymmetric computational cost of the measurement function and the belief state update to achieve a good enough information gain approximation (typically the computing cost of the measurement function is significantly lower). 

Our approach---measurement expectation generation (\our)---as shown in \figref{fig:field_results}, leads to more effective planning decisions. PIM leads to \textit{poor planning decisions} as we explain later in \figref{fig:pims_example} while MC, even with 8 samples and taking \speedup{} longer for a planning iteration compared to \our{}, a seen in \figref{fig:object_follow_runtime}, is not an adequate approximation.
Through a theoretical analysis and a series of extensive experiments, we show our method achieves overall better performance---\textit{effectiveness of planning decisions} and \textit{computational efficiency}---compared to existing methods. In summary, our key contributions are:

\begin{itemize}
    \item We formulate a new, more effective, yet computationally efficient  method for generating measurements from uncertain future states for planning problems (\secref{sec:proposed_method}).
    \item We prove the proposed method achieves a lower measurement prediction error, and consequently a better information gain approximation, than the previous computationally efficient method
    (\secref{sec:proof}). 
    \item We demonstrate the effectiveness and efficiency of our proposed method in \textit{planning for information gathering scenarios} using extensive simulation and \textit{field experiments} with an aerial robot platform (\secref{sec:exp}).
\end{itemize}

In general, our work contributes to realizing autonomous sensing applications under unavoidable real-world limits on hardware cost, weight and size of unmanned aerial vehicles. 

\section{Background and Related Work}
In this section we formally define the problem and discuss related work.  

\subsection{Problem Definition}

We consider an agent performing mobile object tracking using noisy sensor measurements. At time $t$, the agent moves based on the control input $\mathbf{u}_t \in \mathbb{U}_t$, where $\mathbb{U}_t$ is the set of admissible actions. It then receives $m$ noisy measurements $\mathbf{Z}_t = \{ \mathbf{z}_t^{(1)}, \dots, \mathbf{z}_t^{(m)} \}$, composed of possible detections from $n$ objects with states $\mathbf{X}_t = \{ \mathbf{x}^{(1)}_t, \dots, \mathbf{x}^{(n)}_t\}$. Each object produces at most one measurement, and the sensor can produce spurious \textit{clutter} measurements. The object measurements are sensor dependent and modeled by the measurement function, \(\mathbf{z} = h(\mathbf{x})\) while the object dynamics are modeled by the uncertain motion model, \(p(\mathbf{X}_{t+1} | \mathbf{u}_{t+1}, \mathbf{X}_t)\).

Bayesian object tracking algorithms~\cite{mahler_statistical_2007} use \textit{prediction} and \textit{update} procedures to maintain a multi-object probability density \(p(\mathbf{X}_t | \mathbf{u}_{1:t}, \mathbf{Z}_{1:t})\). The motion model is used to predict the density based on the control action, \(\mathbf{u}_{t+1}\), to obtain \(p(\mathbf{X}_{t+1} | \mathbf{u}_{1:t+1}, \mathbf{Z}_{1:t})\). The measurement model is used to update the density with new measurements, \(\mathbf{Z}_{t+1}\), to obtain \(p(\mathbf{X}_{t+1} | \mathbf{u}_{1:t+1}, \mathbf{Z}_{1:t+1})\).
The agent's  objective is to choose the next action $\mathbf{u}^*_{t+1}$ that maximises the expected reward $\mathbb{E}\left[ R(p(\mathbf{X}_t | \mathbf{u}_{1:t}, \mathbf{Z}_{1:t}), \mathbf{u}_{t+1:t+H}, \mathbf{Z}_{t+1:t+H})\right]$ over the finite horizon $H$. Next we describe the information theoretic POMDP problem formulation for achieving the objective. 

\vspace{1mm}
\noindent\textit{Online Information Theoretic Planning with POMDP.~}
\label{sec:pomdp}
A POMDP for information theoretic planning comprises the current belief state \(B_t\), the set of admissible actions \(\mathbf{u}_t \in \mathbb{U}_t\), and a reward determined by the action \(r(B_{t-1}, \mathbf{u}_t)\) reward and information gain \(\Psi(B_t, B_{t-1})\). For object tracking problems, the belief state is the object probability density, \(B_t = p(\mathbf{X}_t | \mathbf{u}_{1:t}, \mathbf{Z}_{1:t})\). Given a sequence of actions \(\mathbf{u}_{t+1:t+H}\) and measurements resulting from those actions \(\mathbf{Z}_{t+1:t+H}\), the reward over the horizon starting from belief state \(B_t\) is computed as the discounted sum of immediate rewards.

\begin{equation}
\label{eq:reward_sum}
 \begin{split}
     R(&B_t, \mathbf{u}_{t+1:t+H}, \mathbf{Z}_{t+1:t+H}) = \\ 
     &\sum_{k=1}^H \lambda^{k-1} \left[r(B_{t+k-1}, \mathbf{u}_{t+k}) + \Psi(B_{t+k}, B_{t+k-1})\right]
 \end{split}
\end{equation}

\noindent
where \(0 < \lambda\ \leq 1\) is the discount factor. The future belief states are a result of the future action and measurement:

\begin{equation}
    B_{t+1} = \text{update}(B_t, \mathbf{u}_{t+1}, \mathbf{Z}_{t+1})
\end{equation}

\noindent
The next action is then selected as:

\begin{equation}
    \mathbf{u}^*_{t+1} = \argmax_{\mathbf{u}_{t+1} \in \mathbb{U}_{t+1}} \mathbb{E}\left[R(B_t, \mathbf{u}_{t+1:t+H}, \mathbf{Z}_{t+1:t+H}) \right]
    \label{eq:expected_reward}
\end{equation}

\noindent where the expectation, \(\mathbb{E}\), is taken with respect to the future actions \(\mathbf{u}_{t+1:t+H}\) and measurements \(\mathbf{Z}_{t+1:t+H}\). 

Information theoretic planning uses measures of \textit{information gain} 
to drive the planner to choose the most \textit{informative} actions. Information quantities, such as Shannon Entropy, Kullback-Leibler divergence, and Renyi Divergence, are commonly used in the reward function~\cite{ristic_note_2011, nguyen_trackerbots_2019, cliff_robotic_2018, dressel_hunting_2019, beard_void_2017}. 

Computing the expectation in~\eqref{eq:expected_reward} requires evaluating all possible sequences of actions and \textit{measurements}. Thus, if there are \(A\) possible actions and \(Z\) possible measurements at each future horizon time, the complexity of computing \(\mathbf{u}^*_{t+1}\) is \(O(A^H Z^H)\). 
Monte Carlo Tree Search methods are often used to compute the expectation in~\eqref{eq:expected_reward} by focusing computational effort on promising action-measurement sequences~\cite{sunberg_online_2018,fischer_information_2020,ott_sequential_2023}. If the belief state can be modeled with a Gaussian distribution, iterative Linear Quadratic Regulator~\cite{sun_belief_2021} methods can be employed. A recent method uses simplified belief states to speedup the computation of information gain~\cite{sztyglic_speeding_2022}. However, it is still necessary to consider the many possible future measurements throughout the tree search for solutions. In general, for non-Gaussian/non-linear systems, \textit{the challenge of computing the expectation in} \eqref{eq:expected_reward} \textit{with respect to future measurements remains}, which is the focus of our work.

Without loss of generality, for the challenge of computing the expectation in \eqref{eq:expected_reward}, we consider the problem of future measurement generation for single-object belief state, i.e., \(B_t = p(\mathbf{x}| \mathbf{u}_{1:t}, \mathbf{Z}_{1:t})\). Because, the  multi-object case is simply an extension of the single object problem (measurements can be generated for each object in the multi-object density).

\subsection{Related Work}
\label{sec:measurement_gen}
Two common approaches are to use Monte-Carlo sampling~\cite{beard_void_2017, dressel_hunting_2019, sunberg_online_2018, nguyen_trackerbots_2019, fischer_information_2020, sztyglic_speeding_2022} or the Predicted Ideal Measurement~\cite{ristic_note_2011, cliff_robotic_2018, yilmaz_particle_2023, fei_conservationbots_2023}. The methods are tasked with generating the next likely measurement, \(\mathbf{Z}_k\), from a belief state \(B_k\) for the challenging problem of computing the expectation in \eqref{eq:expected_reward}.

\vspace{2px}
\noindent\textbf{MC Sampling.~} Here, a random state is sampled from the object density at each time step, \(\mathbf{x} \sim B_t\). Next, the noisy measurement function \(h(\mathbf{x})\) is used to generate a measurement from the sampled state, \(\mathbf{z} = h(\mathbf{x})\).

The expectation is approximated by taking the average over \(M\) repetitions of this process. The number of repetitions directly affects the planning complexity. Effectively, the number of measurements at each horizon time becomes \(1\), however, as this is repeated \(M\) times, the planning complexity is \(O(A^H M)\). Under computational constraints it may be impossible to perform enough repetitions to obtain a good approximation, making it infeasible to use this method.

\vspace{3px}
\noindent\textbf{Predicted Ideal Measurement (PIM).~}The general approach aims to create a deterministic generator such that the number of possible measurements \(Z = 1\), reducing the planning complexity to \(O(A^H)\). To generate a measurement with this approach, first the estimated state is computed from the predicted belief state, \(\bar{\mathbf{x}}= \mean{B_t}\). Then, the \textit{noiseless} measurement function \(\bar{h}(\mathbf{x})\) is applied to the estimated state to obtain the generated measurement, \(\mathbf{z} = \bar{h}(\bar{\mathbf{x}})\).

While the method can significantly reduce planning complexity, it relies \textit{solely} on the estimated state to adequately represent the belief state. But, if the transition or measurement function is non-linear, as is the case in many real-world systems, it is possible for the belief state to become multi-modal. Now, the estimated state may poorly represent the belief state and result in unrealistic predicted measurements, leading to sub-optimal decisions. This problem is illustrated in \figref{fig:pims_example}, where a PIM method predicted measurement leads to a confident, yet extremely unlikely belief state. We propose an alternative, described in \secref{sec:proposed_method}. 

\section{Proposed Measurement Prediction Method}
\label{sec:proposed_method}
We focus on the class of problems where the measurement model is approximately unimodal with no assumptions of the modality of the belief state. Notably, although we assume an approximate unimodal measurement model, our theoretical analysis in Section~\ref{sec:proof} demonstrates our approximation always yields a lower prediction error compared to the computationally efficient alternative, PIM.

The key idea for the proposed method, \our{}, is to compute information gain using the expected measurement instead of a predicted future state as in PIM. \figref{fig:pims_example} illustrates our approach and the formulation's ability to better approximate a measurement from a predicted future belief state compared with PIM when the belief state is multi-modal. More specifically, the expected measurement is computed from a set of randomly sampled measurements. 

First, \(M\) states are randomly sampled from the belief state, \(\mathbf{x}^{(i)} \sim B_t\), where \(i \in [1, \dots, M]\). Next, the noisy measurement function is applied to each of the sampled states to obtain \(M\) sampled measurements, \(\mathbf{z}^{(i)} = h(\mathbf{x}^{(i)})\). The predicted future measurement, $\mathbf{z}$, is generated from the expectation of these sampled measurements:

\begin{equation}
    \label{eq:mc_pims}
    \mathbf{z} = \dfrac{1}{M} \sum_{i=1}^M  \mathbf{z}^{(i)}
\end{equation}

The number of samples affects the variance of the generated measurement. If a sufficient number of samples are used, the variance will be sufficiently small but \(Z=1\), hence, only a single measurement simulation per action is needed. 
Even though the method employs measurement sampling, the computational cost of the measurement function is \textit{typically significantly lower than the cost of the belief update step}. Thus, even if many samples are used, we still expect the method to provide significant improvements to run-time performance over the MC method. This is demonstrated in \secref{sec:runtime}. Further, we show analytically in \secref{sec:proof} that our method can achieve a lower measurement prediction error compared with PIM for arbitrary measurement models.

\subsection{Theoretical Analysis}\label{sec:proof}
The theoretical basis behind the \our\ method is to minimise the error of the predicted measurement with respect to the possible future measurements. To compare PIM and \our, we use the Mean Squared Error (MSE) between the generated measurement \(\hat{z}\) and the posterior measurement distribution. The MSE is defined as: 

\begin{equation}
    \label{eq:measurement_mse}
    \mse(\hat{z} | B_{k+1|k}) = \sum_{z \in Z} p(z | B_{k+1|k}) (\hat{z} - z)^2
\end{equation}

\noindent
where \(p(z| B_{k+1|k})\) is the predicted measurement density from the belief state \(B_{k|k}\). 

\begin{lemma}
The MSE of a PIM measurement is lower bounded by the expected MSE of an \our\ measurement, within some bound \(\varepsilon\) that depends on the number of samples used in \our:
\begin{equation}
\label{eq:pims_geq_mc_pims}
    \mse_{\text{PIM}} + \varepsilon \geq \mean{\mse_{\text{\ourplain}}},~\text{where}~\varepsilon = \dfrac{1}{M} \sigma^2
\end{equation}

\noindent
and \(\sigma^2\) is the variance of \(p(z | B_{k+1|k})\). 
\end{lemma}

\noindent\textit{Proof:}
First, the measurement that minimises \eqref{eq:measurement_mse} is the expected value of the predicted measurement density: 

\begin{equation}
    \bar{z} = \mathbb{E}[z | B_{k+1|k}] = \sum_{z \in Z} p(z | B_{k+1|k}) z
\end{equation}

\noindent which has the following MSE (with the dependence on \(B_{k+1|k}\) omitted for notational simplicity):

\begin{equation}
\begin{split}
    \mse(\bar{z}) &= \sum_{z \in Z} p(z) (\bar{z} - z)^2 \\
    &= \sum_{z \in Z} p(z) \bar{z}^2 - 2 \bar{z} \sum_{z \in Z} p(z) z + \sum_{z \in Z} p(z) z^2 \\ 
    &= \bar{z}^2 - 2 \bar{z}^2 + \mean{z^2} \\
    &= -\bar{z}^2 + \var{z} + \bar{z}^2 = \sigma^2 \\
\end{split}
\end{equation}

\noindent
Note that variance is defined as:

\begin{equation}
    \label{eq:var_eq}
    \var{z} = \mean{z^2} - \mean{z}^2
\end{equation}

Let \(\hat{z}\) be the \our\ measurement \(\mathbf{z}\) from \eqref{eq:mc_pims}. We can see that each sample is selected proportional to \(p(z | B_{k+1|k})\) as \(\mathbf{x}^{(i)} \sim B_{k+1|k}\) and \(h(\mathbf{x}^{(i)})\) generates a measurement proportional to \(p(\mathbf{z} | \mathbf{x}^{(i)})\). Thus, as \(M\) in \eqref{eq:mc_pims} increases, \(\hat{z}\) approaches \(\bar{z}\). However, with only a finite number of samples, \(\hat{z}\) is randomly distributed. According to the Central Limit theorem, the distribution of \(\hat{z}\) is approximately normally distributed around the true mean \(\bar{z}\) with variance \(\varepsilon\) when \(M\) is large enough, \(\hat{z} \sim \mathcal{N}(\bar{z}, \varepsilon)\).

Using this approximation, for a given \(\hat{z}\), the MSE can be computed from \(\bar{z}\), using \eqref{eq:measurement_mse} 

\begin{equation}
    \label{eq:init_mc_pims_mse}
    \mse(\hat{z}) = \mse(\bar{z} + v) = \sum_{z \in Z} p(z)(\bar{z} + v - z)^2
\end{equation}

\noindent
where \(v \sim \mathcal{N}(0, \varepsilon)\) is the approximately normally distributed random noise of the \our\ measurement. Re-writing \eqref{eq:init_mc_pims_mse} by expanding the square gives us:

\begin{equation}
\begin{split}
    \mse(\hat{z}) = &\sum_{z \in Z} \left[ p(z)(\bar{z} - z)^2 \right] \\ &+ 2v \sum_{z \in Z} \left[ p(z)(\bar{z} - z) \right] + v^2 
\end{split}
\end{equation}

Here, the first term on the right hand side is \(\mse(\bar{z})\). The second term is the mean error (ME) of \(\bar{z}\) multiplied by a constant, which is always zero: 

\begin{align}
    \begin{split}
        \me(\bar{z}) &= \sum_{z \in Z}  p(z) (\bar{z} - z) \\
        &= \sum_{z \in Z} p(z) \bar{z} - \sum_{z \in Z} p(z) z = \bar{z} - \bar{z} = 0
    \end{split}
\end{align}

Therefore, the MSE of an \our\ measurement is:

\begin{equation}
    \mse(\hat{z}) = \mse(\bar{z}) + v^2
\end{equation}

\noindent
and the expected MSE is:

\begin{equation}
\label{eq:mc_pims_mse}
\begin{split}
    \mean{\mse(\hat{z})} &= \mse(\bar{z}) + \mean{v^2} \\
    &= \mse(\bar{z}) + \varepsilon
\end{split}
\end{equation}

\noindent
where \(\mathbb{E}[v^2] = \varepsilon\) when \(v \sim \mathcal{N}(0, \varepsilon)\) using \eqref{eq:var_eq}.

Conversely, generating a measurement from the expected object state, as done with the PIM method, produces the following MSE:

\begin{align}
\label{eq:pims_mse}
\begin{split}
    \mse(\bar{h}(\bar{x})) &= \sum_{z \in Z} p(z) (\bar{h}(\bar{x}) - z)^2 \\
    &= \bar{h}(\bar{x})^2 - 2 \bar{z} \bar{h}(\bar{x}) + \mathbb{E}[z^2] \\
    &= \bar{h}(\bar{x})^2 - 2 \bar{z} \bar{h}(\bar{x}) + (\bar{z})^2 + \text{Var}[z] \\
    &= (\bar{h}(\bar{x}) - \bar{z})^2 + \mse(\bar{z})
\end{split}
\end{align}

Here, the first term is always positive or zero, thus, the MSE of the PIM method must be lower bounded by \(\mse(\bar{z})\):

\begin{equation}
    \mse_{\text{PIM}} \geq \mse(\bar{z}).
\end{equation}
Adding $\varepsilon$ to both sides thus satisfies \eqref{eq:pims_geq_mc_pims}, then $\mse_{\text{PIMS}} + \varepsilon \geq \mse(\bar{z}) + \varepsilon$. Hence:
\begin{equation}
    \begin{split}
        \mse_{\text{PIM}} + \varepsilon &\geq \mathbb{E}\left[\mse_{\text{\ourplain}}\right]~~~\blacksquare
    \end{split}
\end{equation}

\vspace{2mm}
\noindent\textbf{Remark 1.~}From \eqref{eq:mc_pims_mse}, our \our\ method's error depends on the predicted measurement variance and can be reduced by increasing samples. But, from \eqref{eq:pims_mse}, the error of the PIM method depends on \(\bar{h}(\cdot)\) and \(\bar{x}\) and cannot be reduced. 

\vspace{2mm}
\noindent\textbf{Remark 2.~}If the system is linear that \(\bar{h}(\bar{x}) = \bar{z}\) and \(\mse_{\text{PIM}} = \mse(\bar{z})\), otherwise, it is possible that \((\bar{h}(\bar{x}) - \bar{z})^2 > 0\). In these cases, there will always be a finite number of samples $M$ for \our, such that $\mathbb{E}[\mse_{\text{\ourplain}}]<\mse_{\text{PIM}}$:
\begin{equation}
    \begin{split}
        \mse(\bar{z}) + (\bar{h}(\bar{x}) - \bar{z})^2 &> \mse(\bar{z}) + \dfrac{1}{M}\sigma^2 \\
        M &> \dfrac{\sigma^2}{(\bar{h}(\bar{x}) - \bar{z})^2}
    \end{split}
\end{equation}

Importantly, this indicates, for any non-linear system, \our{} can perform equally or better than PIM with respect to measurement prediction error.

\section{Simulation and Field Experiments}\label{sec:exp}
We perform a series of extensive simulations and a real-world experiment to evaluate the proposed measurement generation method. In particular:

\begin{itemize}
    \item Object following scenario (\secref{sec:object_following}).
    \item Multi-object tracking \& localisation scenario (\secref{sec:multi_object_localise}).
    \item Run-time analysis of methods (\secref{sec:runtime}).
    \item Field trials with an aerial robot (\secref{sec:field}).
\end{itemize}

In the these experiments, the agent is a quad-copter flying \(50\)~m above ground, with a  \(15\)~m/s top speed, and \(4.0\)~m/s\(^2\) acceleration. For object tracking, we do not solve the object-to-measurement association problem and, as in typical radio source localization problems, assume this can be determined from the measurement~\cite{cliff_robotic_2018,nguyen_trackerbots_2019}. Hence, for each object, we employ a recursive Bayesian filter, in particular a particle-based Bernoulli Filter~\cite{ristic_tutorial_2013}, together with the following measurement and motion (dynamic) models for objects. 

\vspace{2mm}
\noindent\textit{Measurement Model.~}Our experiments consider radio source signal strength measurements periodically emitted from objects (such as rescue beacons). The model follows propagation path loss from the Friis transmission equation, where the received power in dB at a receiver \(u\) is modeled by:
\begin{equation}\label{eq:range-measurement}
    P(x, u) = P^{d_0}_r - 10 n \log_{10} \left( \| x - u \| / d_0 \right) + g_{dir}(x, u) 
\end{equation}
\noindent where \(x\) is the radio source location, \(\| x - u \|\) is the distance from the source to the receiver; \(P^{d_0}_r\) is the received power at some reference distance \(d_0\); \(n\) is the path loss exponent that models the environmental factors; \(g_{dir}(x, u)\) is the antenna gain based on the relative source location and receiver heading. The measurement function is thus:
\begin{equation}
    h(x, u) = P(x, u) + \zeta
    \label{eq:rssi_model}
\end{equation}
\noindent where \(\zeta \sim \mathcal{N}(0, R)\) is the zero-mean, Gaussian measurement noise with variance \(R\). For these experiments, we set \(R=2.5^2\), \(P^{d_0}_r = -7.0\), \(d_0 = 1.0\), \(n = 1.5\) from field data, and \(g_{dir}(x, u)\) to use a monopole antenna gain pattern. We use this measurement model to demonstrate how a non-linear measurement model can affect the planning decisions.

\vspace{2mm}
\noindent\textit{Motion Models.~}
In the simulated experiments, we use four different object motion models: i)~Random Walk (RW), ii)~Constant Velocity (CV), iii)~CV with infrequent turning (CV-IFT), and iv)~CV-IFT with mismatched filter. To model a maneuvering object, the CV-IFT model uses an interacting multiple model (IMM) where the object state switches between three modes; 1) CV; 2) constant turn rate with acceleration (CTRA)~\cite{svensson_derivation_2019}; 3) negative CTRA to allow the object to move straight, turn left, and turn right. For this case, we use the IMM particle filter~\cite{mcginnity_multiple_2000} for state estimation. All of the models are known to the filter, except the CV-IFT with mismatched filter scenarios; here, the filter assumes a CV model for objects but with increased noise to compensate for possible manoeuvres. This \textit{hard setting} simulates a dynamic model mismatch experienced in practical settings.

\vspace{2mm}

\noindent\textit{Path Planning.~}
The action set for the agent consists of the eight cardinal directions, where the agent will simply head in the selected direction until a new direction is chosen by the path planning algorithm.
Online path planning is implemented by approximating the expectations in \eqref{eq:expected_reward} with simulations.
For each action \(\mathbf{u}_{t+1}\), action sequences are simulated starting with \(\mathbf{u}_{t+1}\), \(\hat{\mathbf{u}}_{t+1:t+H} = (\mathbf{u}_{t+1}, \dots, \mathbf{u}_{t+H})\). The simulation is then played out by repeating the following four-step process:

\vspace{2mm}
\begin{enumerate}[itemsep=1pt,parsep=3pt,topsep=0pt]

    \item Use the next action \(\mathbf{u}_{k+1}\) to compute the predicted belief state $B_{k+1|k} = p(\mathbf{X}_{k+1} | \mathbf{u}_{1:k+1}, \mathbf{Z}_{1:k})$. 

    \item Generate the measurements \(\mathbf{Z}_{k+1}\) from the predicted belief state using one of the measurement prediction methods: $\mathbf{Z}_{k+1} = g(B_{k+1|k})$ \label{step:measurement}

    \item Update the belief state using the generated measurement to obtain the posterior belief state:\\ \(B_{k+1} = p(\mathbf{X}_{k+1} | \mathbf{u}_{1:k+1}, \mathbf{Z}_{1:k+1})\)

    \item Compute the reward of the action:\\ $r_k = r(B_k, \mathbf{u}_{k+1}) + \Psi(B_{k+1}, B_k)$.
\end{enumerate}

\vspace{2mm}

Here, \(r_k\) is used to compute the sum of discounted rewards \eqref{eq:reward_sum}, and the action with the highest average sum is selected. We choose to use Renyi Divergence to formulate our reward as it is readily computed from a particle approximation of a belief density and is successfully used in variety of scenarios~\cite{nguyen_trackerbots_2019, ristic_note_2011,beard_void_2017}. Renyi Divergence between the two distributions \(p\) and \(q\) is computed as:

\begin{equation}\label{eq:renyidiv}
    \Psi(p, q) = \dfrac{1}{\alpha - 1} \log \int p(x)^{\alpha} q(x)^{1-\alpha} \, dx
\end{equation}

\noindent 
where \(\alpha > 0\) controls the effect of the tails of the two distributions. We set \(\alpha = 0.1\) for all experiments as this has demonstrated the best planning performance, although the choice does not greatly impact performance.

\subsection{Object Following}
\label{sec:object_following}
Here, we evaluate the impact of the different measurement prediction methods on planning to follow and track an object. In each scenario, a controllable agent aims to minimize the uncertainty of an object. The planner re-plans every \(5\) seconds, with a \(10\) second look-ahead time. Planning is performed using a belief state of \(1000\) particles sampled from the filter. The search space is \(4\)~km by \(4\)~km square, with the object being at a height of \(0\)~m, and the agent starting at the centre. For the MC method, each action sequence is simulated \(8\) times (MC-8), chosen so the simulations complete in a reasonable time frame. For the \our\ method, we use \(512\) measurement samples as negligible improvement is seen with more.  

\begin{figure}[!ht]
\vspace{-3mm}
    \centering
    \subfloat{\includegraphics[width=0.28\linewidth]{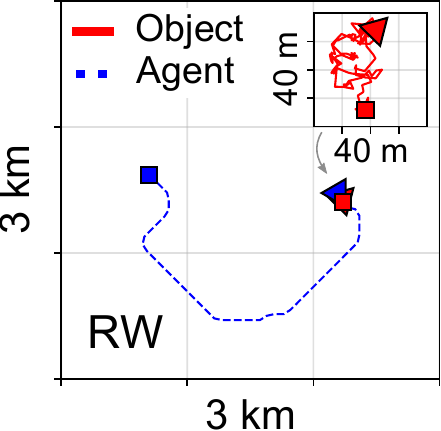}\label{fig:follow_static_visualise}}
    \hspace{0.03\linewidth}
    \subfloat{\includegraphics[width=0.28\linewidth]{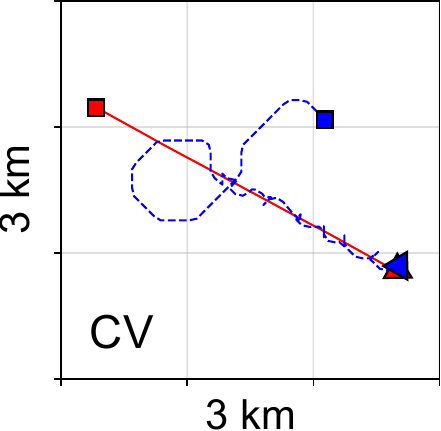}\label{fig:follow_cv_visualise}}
    \hspace{0.03\linewidth}
    \subfloat{\includegraphics[width=0.28\linewidth]{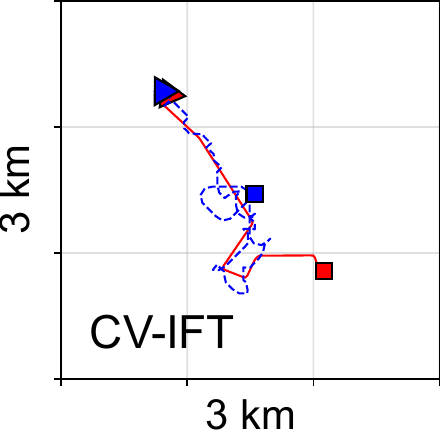}\label{fig:follow_mmift_visualise}}
    \caption{Agent and object path extracts in the object following experiments ($\square$: represent the starting positions; $\triangle$: represent the ending positions).}
    \label{fig:follow_example}
    \vspace{-3mm}
\end{figure}

\vspace{2mm}
\noindent\textit{Trajectory Generation.~}For each scenario, we perform extensive Monte-Carlo (MC) simulations. First, we generate \(50\) unique starting states for each scenario. Then, we run \(30\) MC simulations for each starting state, giving the planner \(600\) simulated seconds to find and follow the object.
Overall, \(1500\) simulations are performed for each scenario, totalling to \(250\) hours of simulated flight time. The object states for each unique trajectory are intialized in the following way  (examples of these trajectories are illustrated in \figref{fig:follow_example}):

\begin{itemize}
    \item \textit{Position (all models):} The initial object position for trajectories are uniformly sampled from the search space.
    \item \textit{Velocity (CV and CV-IFT):} The velocities are initialized by uniformly sampling a direction and speed from the intervals \([0^{\circ}, 360^{\circ}]\) and \([5~m/s, 6~m/s]\) respectively.
    \item \textit{Turn Rate (CV-IFT):} The turn rate is uniformly sampled from the interval \([10^{\circ} /s, 15^{\circ} /s]\).
    \item \textit{IMM Transition Matrix (CV-IFT):} The state transitions are set to occur every \(1\) second with a transition probability matrix with the rows, \(c_1 = [0.975, 0.0125, 0.0125]\), \(c_2 = [0.225, 0.775, 0.0]\), \(c_3 = [0.225, 0.0, 0.775]\).
\end{itemize}

\vspace{1mm}
\noindent\textit{Performance Measures.~}After every filter update, we record two metrics; i)~\textit{Error}: the absolute error between the estimated object position and the ground truth position; and ii)~\textit{Tr(Cov)}: the trace of the object positional covariance.

\begin{figure}[!ht]
    \centering
    \subfloat{\includegraphics[width=\linewidth]{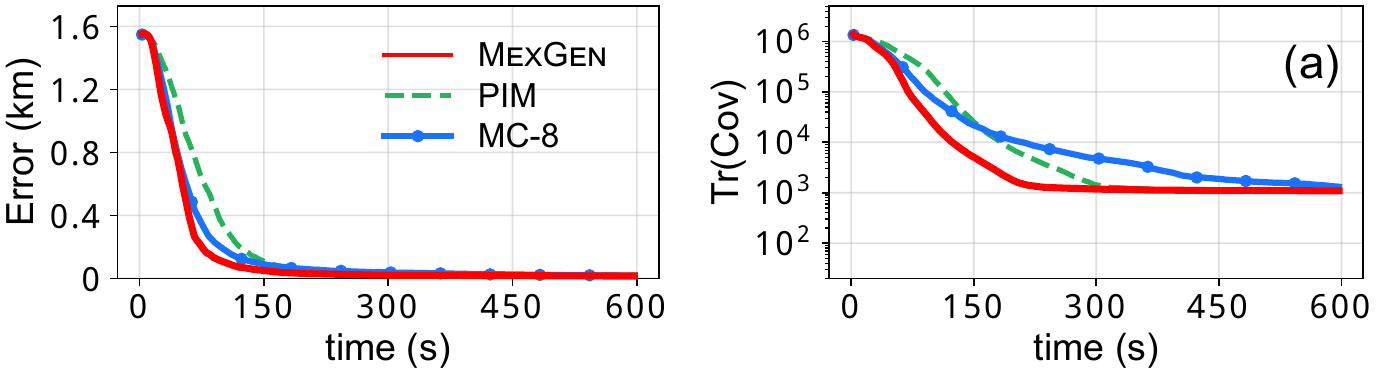}\label{fig:follow_static}} \\
    \subfloat{\includegraphics[width=\linewidth]{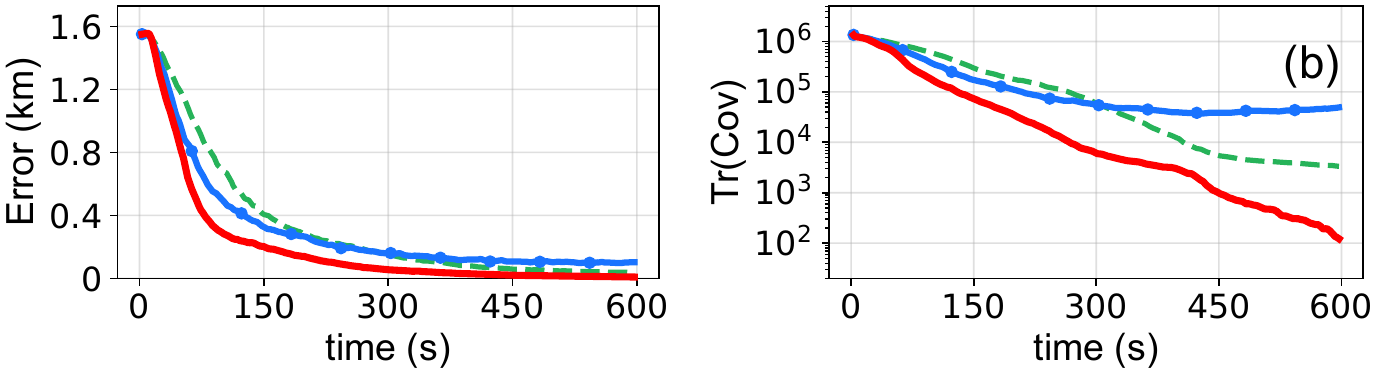}\label{fig:follow_cv}} \\
    \subfloat{\includegraphics[width=\linewidth]{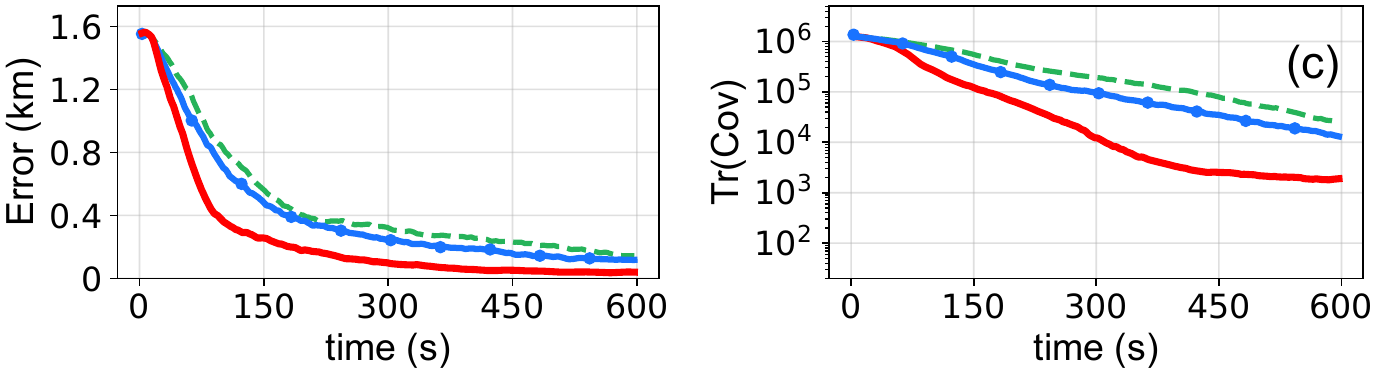}\label{fig:follow_cv_ift}} \\
    \subfloat{\includegraphics[width=\linewidth]{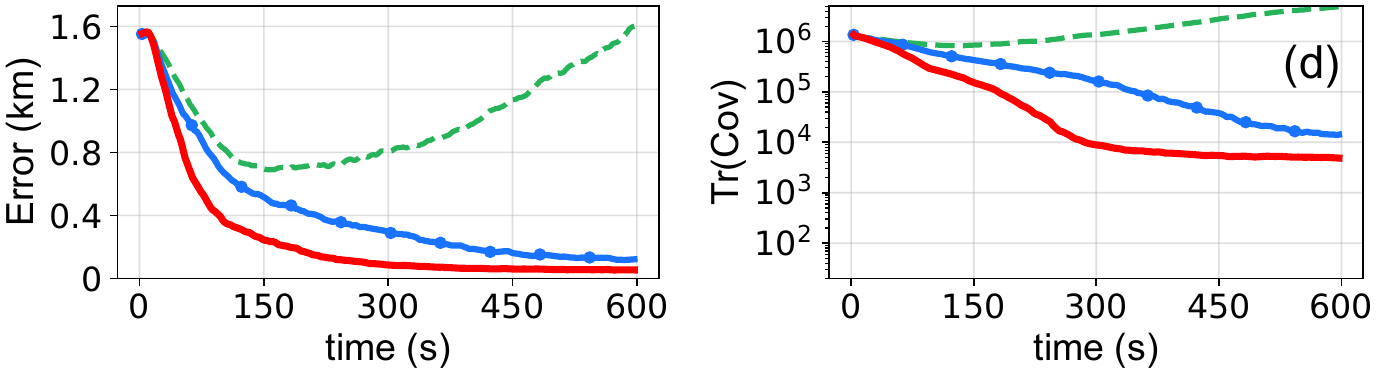}\label{fig:follow_cv_ift_mm}}
    \caption{Object following results showing \textit{estimation error} and \textit{uncertainty of estimates} (position components of the covariance matrix) as Tr(Cov) from 50 unique initial states $\times$~30~MC simulations per method. The median values are plotted at each time step for: \protect\subref{fig:follow_static}~Random Walk, \protect\subref{fig:follow_cv}~Constant Velocity, \protect\subref{fig:follow_cv_ift} CV-IFT, and \protect\subref{fig:follow_cv_ift_mm} CV-IFT with mismatched filter (the \textit{hard} estimation problem setting). More effective planning decisions with our \our{} method yield improved tracking accuracy and lower uncertainty over time.}
    \label{fig:object_following_results}
\end{figure}

\vspace{2mm}
\noindent\textit{Results.~}Importantly, the results in \figref{fig:object_following_results} show that \our\ consistently leads to improved error and lower covariance compared to other methods. Overall, the MC method yields increased variance in planning decisions, often causing the agent to repeatedly change direction. Consequently, making it more difficult for the agent to maintain proximity with the object and generally resulting in higher covariance and error compared to \our. Interestingly, in \figref{fig:follow_cv}, we see the MC and PIM methods display high uncertainty over time compared to \our{}. 
This illustrates the impact of poor information gathering actions resulting from the methods. Objects in the CV model can travel away from the agent and poor action choices can increase the object-to-agent distance; consequentially, leading to less informative measurements. Although in \figref{fig:follow_cv_ift}, the impact is less pronounced, as CV-IFT model keeps the object closer to its initial position because the object frequently changes direction (see \figref{fig:follow_example}). 

We can observe a more prominent impact of poor information gathering action decisions with PIM under the CV-IFT with mismatched filter settings. \figref{fig:follow_cv_ift_mm} shows that the agent using PIM is unable to find the object with the error and covariance increasing over time. The mismatch between the object dynamics and the filter's motion model, reflective of settings in practice, becomes especially problematic for PIM-based measurement predictions. Although increased process noise could account for the mismatch by allowing the belief state to become more uncertain, the point estimate fails to capture the mode of the belief state distribution. Consequently, the PIM method predicts future measurements poorly and leads to selecting ineffective information gathering actions.

\subsection{Multi-Object Tracking and Localisation}
\label{sec:multi_object_localise}

In this section, we evaluate the impact of the measurement prediction methods on planning time to localize \(4\) objects. This considers an information gathering scenario to find and inspect a number of objects, e.g., to monitor the health of endangered animals~\cite{cliff_robotic_2018,nguyen_trackerbots_2019}. Here, planning is performed by opting to reduce the position uncertainty of the closest object---the action that maximizes information gain---as determined by the estimated states. 
When the trace of position components of the covariance for an object indicates uncertainty is less than approximately \(50\)~m with 95\% confidence, it is considered localized and the agent will not select it again for planning. 

In these experiments, we reduce the initial search space to a 1~km by 1~km area to ensure the MC simulations complete within a reasonable time-frame. But, we increase the look-ahead time to 20~s to better plan for events where two objects are greatly separated as they move in opposite directions. 

\begin{figure}[!t]
    \centering
    \subfloat{\includegraphics[width=0.28\linewidth]{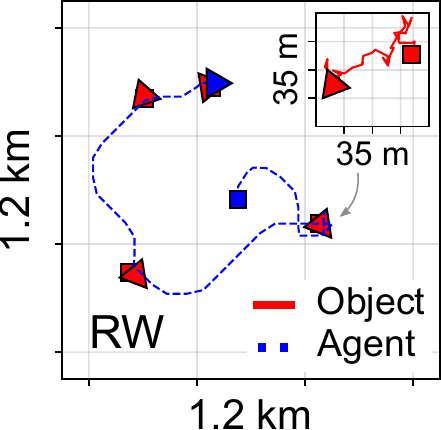}\label{fig:static_visualise}}
    \hspace{0.03\linewidth}
    \subfloat{\includegraphics[width=0.28\linewidth]{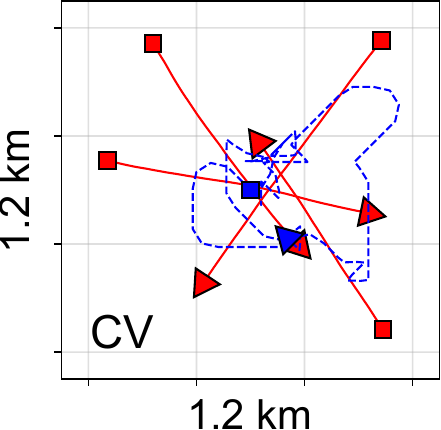}\label{fig:cv_visualise}}
    \hspace{0.03\linewidth}
    \subfloat{\includegraphics[width=0.28\linewidth]{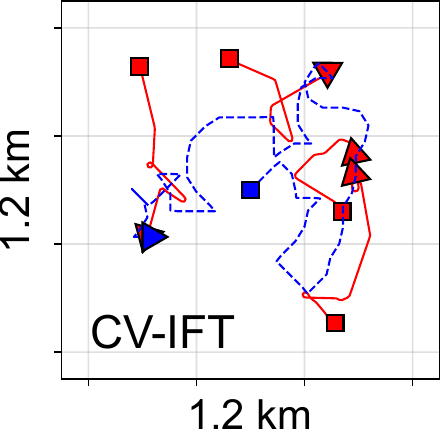}\label{fig:mmift_visualise}}
    \caption{Agent and object path extracts in the multi-object tracking and localisation experiments ($\square$: the starting positions; $\triangle$: ending positions).}
    \label{fig:localise_example}
    \vspace{-3mm}
\end{figure}

\noindent\textit{Object Trajectory Generation.~}In these experiments, we used the trajectory generation method in \secref{sec:object_following} but reduce the speed of the objects of the CV and CV-IFT case to be sampled from the interval \([2.5\)~m/s,~\(3\)~m/s\(]\) and select starting positions such that each object remains within the search area for at least \(650\) seconds. Example trajectories for the experiments are given in \figref{fig:localise_example}. Again, we generate results from 50 unique initial states $\times$ 30 Monte Carlo simulations (1500 MC simulations) conducted per method per motion model.

\vspace{2mm}
\noindent\textit{Performance Measures.~} i)~\textit{Error}: we record the absolute estimation error of objects when they are considered localised; ii)~\textit{Time}: once all objects are localised, we record the time for task completion. If the time taken $>1000$~s we consider the task a fail and discard the result; and iii)~\textit{Success}: report the ratio of successful trials to total runs as success rate. 

\vspace{2mm}
\noindent\textit{Results.~}The mean and standard deviation results presented in \tableref{tab:mt_localise} show that \our\ is able to localise the objects on average faster than the other two methods. Similar to the results in \secref{sec:object_following}, we see a greater difference in the results as the complexity of the object motion increases. Further, we see a drastic decrease in the success rate of MC and PIM in the CV-IFT and CV-IFT filter mismatch cases, while \our\ remains close to \(1.0\). In the \textit{Mismatch} filter case, the hard setting, the success rate of PIM and MC are reduced further, while \our\ remains high. Consequently, we can observe \our\ to provide better measurement approximations under high model and state uncertainty leading to improved selections of information gathering actions.  

\begin{table}[t!]
\centering
\setlength{\tabcolsep}{4.5pt}
\caption{Multi-Object tracking \& localisation results from 50 unique initial states $\times$~30 MC simulations per method per model.
}
\label{tab:mt_localise}
\begin{tabular}{c l c c c}
\toprule
Model & Method & Success & Error \(\pm \sigma\)~(m) & Time \(\pm \sigma\)~(s)\\
\toprule
& PIM & 1.0 & \(12.4 \pm 9.2\) & \(297.7 \pm 93.7\) \\  
\textbf{RW} & MC-8 & 1.0 & \(12.4 \pm 8.6\) & \(255.6 \pm 64.4\) \\
& \ourplain & 1.0 & \(12.1 \pm 8.4\) & \(\boldsymbol{231.0 \pm 55.6}\) \\
\midrule
& PIM & 0.98 & \(11.7 \pm 9.2\) & \(444.7 \pm 115.1\) \\  
\textbf{CV} & MC-8 & 1.0 & \(11.7 \pm 8.9\) & \(405.2 \pm 92.0\) \\
& \ourplain & 1.0 & \(12.1 \pm 9.7\) & \(\boldsymbol{360.1 \pm 74.8}\) \\
\midrule
& PIM & 0.53 & \(12.8 \pm 9.7\) & \(645.5 \pm 194.7\) \\  
\textbf{CV-IFT} & MC-8  & 0.90 & \(12.5 \pm 9.4\) & \(558.6 \pm 183.6\) \\
& \ourplain & \textbf{0.99} & \(12.6 \pm 9.6\) & \(\boldsymbol{465.5 \pm 159.9}\) \\
\midrule
\multirow{3}{*}{\shortstack{\textbf{CV-IFT}\\\textbf{Mismatch}\\(hard setting)}} & PIM & 0.22 & \(12.0 \pm 7.4\) & \(726.4 \pm 181.4\) \\
& MC-8  & 0.66 & \(11.5 \pm 7.1\) & \(651.0 \pm 185.2\) \\
& \ourplain & \textbf{0.90} & \(11.5 \pm 7.1\) & \(\boldsymbol{567.4 \pm 185.9}\) \\
\bottomrule
\end{tabular}
\vspace{-5mm}
\end{table}

\subsection{Run-Time Analysis}
\label{sec:runtime}

\begin{figure}[!ht]
    \centering
    \includegraphics[width=\linewidth]{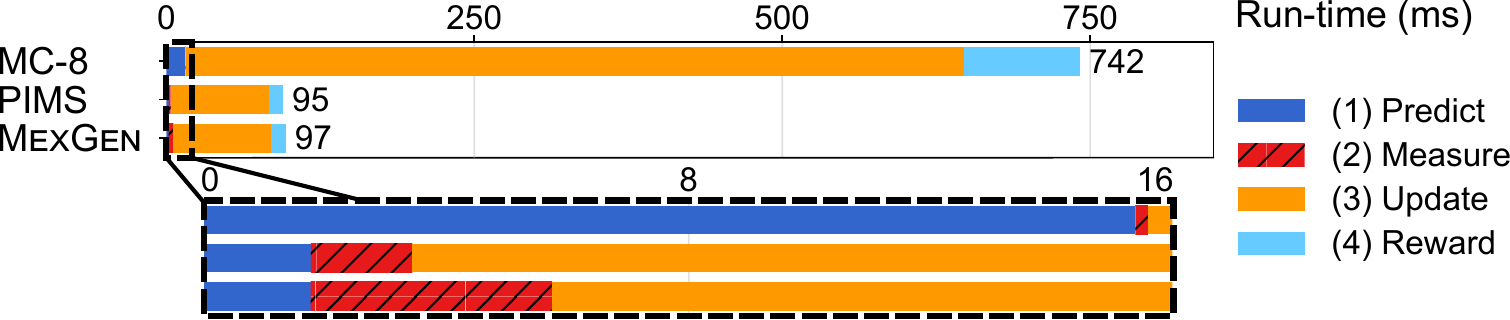}
    \caption{Run-time analysis results for planners with the different measurement prediction methods running on the processor used on our aerial robot (total time is divided into the planning stages in \secref{sec:exp}). Our \our{} method achieves \speedup{} speed-up compared to MC-8 and significantly better task performance than PIM (see Table~\ref{tab:mt_localise} and \ref{tab:field_results}). Notably, the relatively small measure time of MC-8 is because PIM and \our{} measure times include the cost of the expectation. The results demonstrate the asymmetric computational cost of the measurement function (\textit{Measure} time) and the belief state update (\textit{Update} time) we leverage in our approximation.}
    \label{fig:object_follow_runtime}
    \vspace{-3mm}
\end{figure}

In this section we analyse the run-time of the planner with the different measurement generation methods. All experiments are run on a DJI Manifold 2-C processor used on our aerial robot. To provide greater insights into how the measurement generation method affects planning time, we record the time spent in each step of the planning algorithm (\secref{sec:pomdp}) separately. The medians of these results are presented in \figref{fig:object_follow_runtime}, with the sum of medians approximating the expected total run-time.

From \figref{fig:object_follow_runtime}, we see that the belief update step dominates the run-time for all methods. Importantly, generating the measurement samples in \our\ only results in a small increase compared to PIM, while using MC with \(8\) sequence repetitions results in a significant increase to run-time. Notably, computational costs are exacerbated when the state space (information to gather) is large, where predicting and updating the belief state can become computationally taxing as the probability of being in each of the many states must be maintained. Consequently, the computational cost of an already complex problem becomes prohibitive when a large horizon time is considered without an effective approximation, such as our \our.

\begin{figure}[b]
    \centering
    \subfloat{\includegraphics[width=\linewidth]{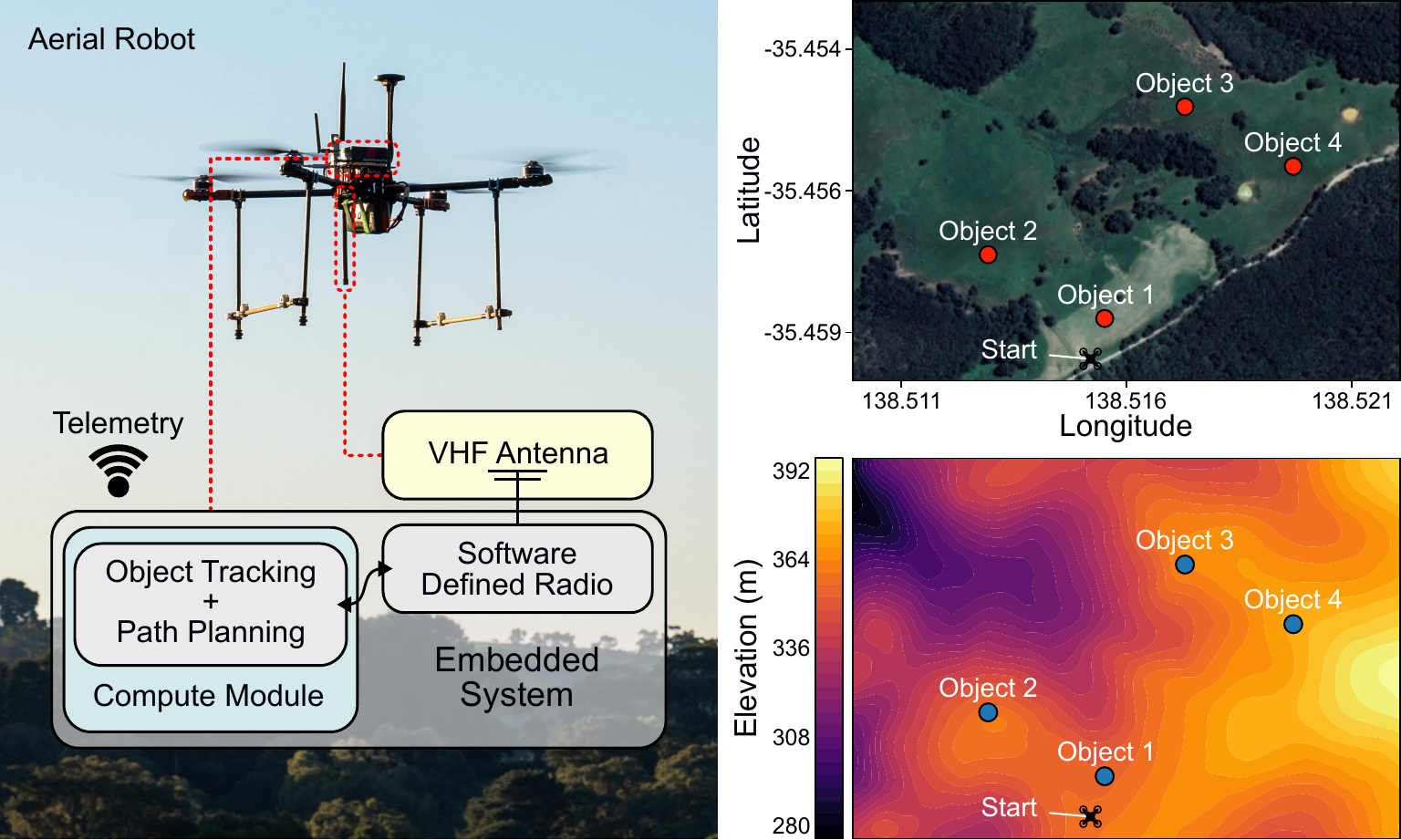}\label{fig:field_setup}} 
    \caption{The aerial robot with the on-board sensor systems for detecting the Very High Frequency (VHF) radio sources used in our field trials and satellite image of the 41 hectare site with a hilly terrain used for the trials.}
    \label{fig:victor}
\end{figure}

\begin{figure*}[t]
    \centering
    \includegraphics[width=0.8\linewidth]{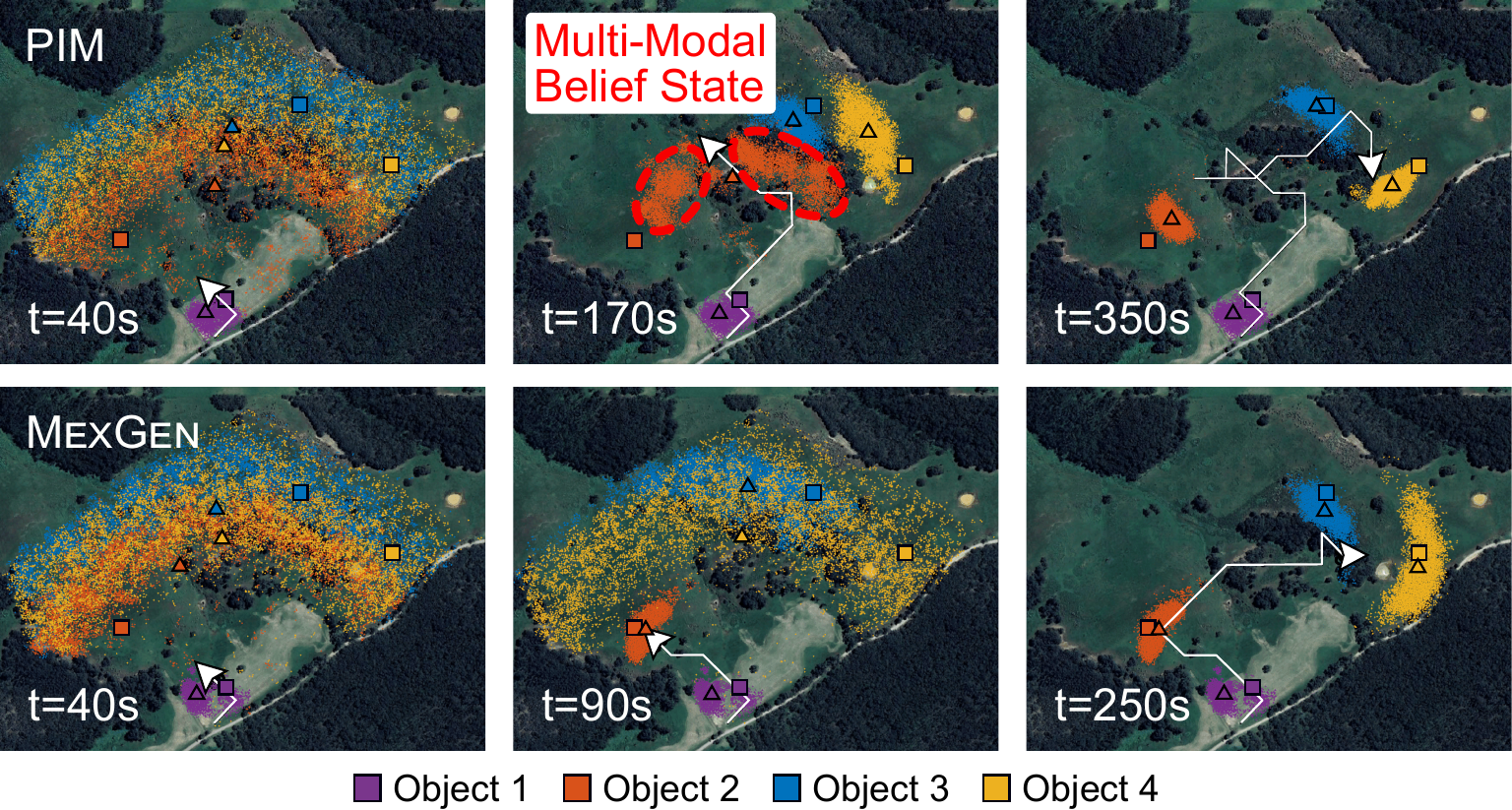}
    \caption{Example paths followed using PIM~vs.~\ourplain~(ours). PIM leads to selecting less effective actions as it can overestimate the information gain for actions. For example, at $t=170$, PIM guided robot heads towards the center of the multi-modal belief state, away from the true object location and conducts multiple maneuvers near the center, seen at \(t=350\). In contrast, our approach leads to selecting more effective actions as it yields better estimates of information gain for actions; the robot maneuvers closer to the objects of interest to reduce range measurement errors and estimation uncertainty.}
    \label{fig:field_examples}
\end{figure*}

\subsection{Field Experiments}
\label{sec:field}

In this section we evaluate the effectiveness of our measurement generation method in a real-world object localisation scenario using a lightweight quad-copter platform. In this scenario---modeled after applications in~\cite{dressel_pseudo-bearing_2018,dressel_hunting_2019,nguyen_trackerbots_2019,cliff_robotic_2018,bayram_gathering_2016}, using the measurement model in~\eqref{eq:range-measurement}---four radio tags (Objects 1-4) pulsing every second on unique frequencies  are placed within a hilly environment. The quad-copter and the hilly environment are depicted in~\figref{fig:victor}. The system can run fully autonomously. 
Planning proceeds as in~\secref{sec:multi_object_localise}. 
Importantly, due to the challenging terrain and to enable us to compare across methods, the tags are stationary throughout the experiments. The starting location is the entrance to the field site, where our equipment was located and with a safe take-off and landing zone for the drone.

\vspace{2mm}
\noindent\textit{Results.~}The localisation time and error results are in \tableref{tab:field_results} with two example missions shown in \figref{fig:field_examples} with heat-maps of the paths for all missions presented in \figref{fig:field_results}. \our\ was able to consistently localise the objects in shorter times than PIM and MC-8 while all methods achieve similar localisation accuracy. \figref{fig:field_results} indicates that PIM tends to head towards the centre of the environment where the initial radio source point estimates are, while \our\ selects the more effective, direct paths, towards the objects. \figref{fig:field_examples} illustrates an instance where PIM generates erroneous measurements that over estimate the information gain and result in less effective actions. Even with a \speedup{} increase in computational time (see \figref{fig:object_follow_runtime}), MC-8 demonstrated poor action selection and longer localisation times compared to \our. These results confirm our findings with simulation-based experiments.

\begin{table}[t!]
\centering
\caption{Field experiment results with our aerial robot platform.}
\label{tab:field_results}
\begin{tabular}{ c c c c }
\toprule
Method & Missions & Error \(\pm \sigma\) (m) & Time \(\pm \sigma\) (s)\\ 
\midrule
\ourplain\ (\textbf{Ours}) & 10 & \(68 \pm 16\) & \(\mathbf{290 \pm 46}\) \\
MC-8 & 10 & \(75 \pm 25\) & \(326 \pm 45\) \\  
PIM & 10 & \(69 \pm 12\) & \(333 \pm 40\) \\  
\bottomrule
\end{tabular}
\end{table}

\section{Conclusions}
We consider the problem of estimating the information gain of future actions with respect to the large space of possible future measurements. 
We propose a new, computationally efficient and effective method for predicting measurements to approximate information gain for  online planning algorithms to select the best information gathering actions. We demonstrated our method---\our---provides improved planning performance compared to baseline methods, with a low computational complexity in a variety of simulated and real-world scenarios. In general, the belief state and measurement function can be multi-modal, requiring a computationally expensive method considering multiple measurements, such as MC, to plan effectively. We have not considered such a setting. However, even in these settings, \our{} will always yield a lower prediction error compared to PIM (see \secref{sec:proof}).

\bibliographystyle{IEEEtran}
\bibliography{IEEEabrv,bibliography}

\begin{thebibliography}{10}
\providecommand{\url}[1]{#1}
\csname url@rmstyle\endcsname
\providecommand{\newblock}{\relax}
\providecommand{\bibinfo}[2]{#2}
\providecommand\BIBentrySTDinterwordspacing{\spaceskip=0pt\relax}
\providecommand\BIBentryALTinterwordstretchfactor{4}
\providecommand\BIBentryALTinterwordspacing{\spaceskip=\fontdimen2\font plus
\BIBentryALTinterwordstretchfactor\fontdimen3\font minus \fontdimen4\font\relax}
\providecommand\BIBforeignlanguage[2]{{%
\expandafter\ifx\csname l@#1\endcsname\relax
\typeout{** WARNING: IEEEtran.bst: No hyphenation pattern has been}%
\typeout{** loaded for the language `#1'. Using the pattern for}%
\typeout{** the default language instead.}%
\else
\language=\csname l@#1\endcsname
\fi
#2}}

\bibitem{schedl_autonomous_2021}
D.~C. Schedl, I.~Kurmi, and O.~Bimber, ``An autonomous drone for search and rescue in forests using airborne optical sectioning,'' \emph{Sci. Robot.}, vol.~6, no.~55, p. eabg1188, 2021.

\bibitem{tao_seer_2023}
Y.~Tao, Y.~Wu, B.~Li, F.~Cladera, A.~Zhou, D.~Thakur, and V.~Kumar, ``{SEER}: {Safe} {Efficient} {Exploration} for {Aerial} {Robots} using {Learning} to {Predict} {Information} {Gain},'' in \emph{{IEEE} Int. Conf. Robot. Autom. (ICRA)}, 2023, pp. 1235--1241.

\bibitem{vander2014}
J.~Vander~Hook, P.~Tokekar, and V.~Isler, ``Cautious greedy strategy for bearing-only active localization: Analysis and field experiments,'' \emph{Journal of Field Robotics}, vol.~31, no.~2, pp. 296--318, 2014.

\bibitem{vander2016}
H.~Bayram, J.~V. Hook, and V.~Isler, ``Gathering bearing data for target localization,'' \emph{IEEE Robotics and Automation Letters}, vol.~1, no.~1, pp. 369--374, 2016.

\bibitem{cliff_robotic_2018}
O.~M. Cliff, D.~L. Saunders, and R.~Fitch, ``Robotic ecology: {Tracking} small dynamic animals with an autonomous aerial vehicle,'' \emph{Sci. Robot.}, vol.~3, no.~23, p. eaat8409, 2018.

\bibitem{fei_conservationbots_2023}
F.~Chen, H.~V. Nguyen, D.~A. Taggart, K.~Falkner, S.~H. Rezatofighi, and D.~C. Ranasinghe, ``{ConservationBots}: Autonomous aerial robot for fast robust wildlife tracking in complex terrains,'' \emph{J. Field Robot.}, 2023.

\bibitem{bayram_gathering_2016}
H.~Bayram, J.~Vander~Hook, and V.~Isler, ``Gathering bearing data for target localization,'' \emph{Robot. and Autom. Letters}, vol.~1, no.~1, pp. 369--374, 2016.

\bibitem{yilmaz_particle_2023}
M.~K. Yılmaz and H.~Bayram, ``Particle filter-based aerial tracking for moving targets,'' \emph{J. Field Robot.}, vol.~40, no.~2, pp. 368--392, 2023.

\bibitem{nguyen_trackerbots_2019}
H.~V. Nguyen, M.~Chesser, L.~P. Koh, S.~H. Rezatofighi, and D.~C. Ranasinghe, ``{TrackerBots}: {Autonomous} unmanned aerial vehicle for real-time localization and tracking of multiple radio-tagged animals,'' \emph{J. Field Robot.}, vol.~36, no.~3, pp. 617--635, 2019.

\bibitem{dressel_hunting_2019}
L.~Dressel and M.~J. Kochenderfer, ``Hunting {Drones} with {Other} {Drones}: {Tracking} a {Moving} {Radio} {Target},'' in \emph{{IEEE} Int. Conf. Robot. Autom. (ICRA)}, 2019, pp. 1905--1912.

\bibitem{araya_pomdp_2010}
M.~Araya, O.~Buffet, V.~Thomas, and F.~Charpillet, ``A {POMDP} {Extension} with {Belief}-dependent {Rewards},'' in \emph{Adv. Neural Inf. Process. Syst. (NIPS)}, 2010.

\bibitem{sunberg_online_2018}
Z.~Sunberg and M.~Kochenderfer, ``Online {Algorithms} for {POMDPs} with {Continuous} {State}, {Action}, and {Observation} {Spaces},'' \emph{Int. Conf. on Autom. Planning and Scheduling}, vol.~28, no.~1, pp. 259--263, 2018.

\bibitem{fischer_information_2020}
J.~Fischer and O.~S. Tas, ``Information {Particle} {Filter} {Tree}: {An} {Online} {Algorithm} for {POMDPs} with {Belief}-{Based} {Rewards} on {Continuous} {Domains},'' in \emph{Int. Conf. Mach. Learning (ICML)}, 2020, pp. 3177--3187.

\bibitem{sztyglic_speeding_2022}
O.~Sztyglic and V.~Indelman, ``Speeding up {POMDP} {Planning} via {Simplification},'' in \emph{{IEEE} Int. Conf. Intell. Robots Syst. (IROS)}, 2022, pp. 7174--7181.

\bibitem{ott_sequential_2023}
J.~Ott, E.~Balaban, and M.~J. Kochenderfer, ``Sequential {Bayesian} {Optimization} for {Adaptive} {Informative} {Path} {Planning} with {Multimodal} {Sensing},'' in \emph{{IEEE} Int. Conf. Robot. Autom. (ICRA)}, 2023, pp. 7894--7901.

\bibitem{burks_optimal_2019}
L.~Burks, I.~Loefgren, and N.~R. Ahmed, ``Optimal continuous state pomdp planning with semantic observations: {A} variational approach,'' \emph{{IEEE} Trans. Robot.}, vol.~35, no.~6, pp. 1488--1507, 2019.

\bibitem{cai_hyp-despot_2021}
P.~Cai, Y.~Luo, D.~Hsu, and W.~S. Lee, ``{HyP}-{DESPOT}: {A} hybrid parallel algorithm for online planning under uncertainty,'' \emph{Int. J. Robot. Res.}, vol.~40, no. 2-3, pp. 558--573, 2021.

\bibitem{Nguyen_Rezatofighi_Vo_Ranasinghe_2020}
H.~V. Nguyen, H.~Rezatofighi, B.-N. Vo, and D.~C. Ranasinghe, ``Multi-objective multi-agent planning for jointly discovering and tracking mobile objects,'' \emph{Proceedings of the AAAI Conference on Artificial Intelligence (AAAI)}, vol.~34, no.~05, pp. 7227--7235, 2020.

\bibitem{papadimitriou_complexity_1987}
C.~H. Papadimitriou and J.~N. Tsitsiklis, ``The complexity of {Markov} decision processes,'' \emph{Mathematics of operations research}, vol.~12, no.~3, pp. 441--450, 1987.

\bibitem{beard_void_2017}
M.~Beard, B.-T. Vo, B.-N. Vo, and S.~Arulampalam, ``Void probabilities and {Cauchy}–{Schwarz} divergence for generalized labeled multi-{Bernoulli} models,'' \emph{{IEEE} Trans. Signal Process.}, vol.~65, no.~19, pp. 5047--5061, 2017.

\bibitem{ristic_note_2011}
B.~Ristic, B.-N. Vo, and D.~Clark, ``A {Note} on the {Reward} {Function} for {PHD} {Filters} with {Sensor} {Control},'' \emph{{IEEE} Trans. Aerosp. Electron. Syst.}, vol.~47, no.~2, pp. 1521--1529, 2011.

\bibitem{mahler_statistical_2007}
R.~P.~S. Mahler, \emph{Statistical {Multisource}-{Multitarget} {Information} {Fusion}}.\hskip 1em plus 0.5em minus 0.4em\relax USA: Artech House, Inc., 2007.

\bibitem{sun_belief_2021}
K.~Sun and V.~Kumar, ``Belief {Space} {Planning} for {Mobile} {Robots} {With} {Range} {Sensors} {Using} {iLQG},'' \emph{Robot. and Autom. Letters}, vol.~6, no.~2, pp. 1902--1909, Apr. 2021.

\bibitem{ristic_tutorial_2013}
B.~Ristic, B.-T. Vo, B.-N. Vo, and A.~Farina, ``A tutorial on {Bernoulli} filters: theory, implementation and applications,'' \emph{{IEEE} Trans. Signal Process.}, vol.~61, no.~13, pp. 3406--3430, 2013.

\bibitem{svensson_derivation_2019}
D.~Svensson, ``Derivation of the discrete-time constant turn rate and acceleration motion model,'' in \emph{{Sensor} {Data} {Fusion}: {Trends}, {Solutions}, {Applications} ({SDF})}, 2019, pp. 1--5.

\bibitem{mcginnity_multiple_2000}
S.~McGinnity and G.~Irwin, ``Multiple model bootstrap filter for maneuvering target tracking,'' \emph{{IEEE} Trans. Aerosp. Electron. Syst.}, vol.~36, no.~3, pp. 1006--1012, 2000.

\bibitem{dressel_pseudo-bearing_2018}
L.~Dressel and M.~J. Kochenderfer, ``Pseudo-bearing {Measurements} for {Improved} {Localization} of {Radio} {Sources} with {Multirotor} {UAVs},'' in \emph{{IEEE} Int. Conf. Robot. Autom. (ICRA)}, 2018, pp. 6560--6565.

\end{thebibliography}

\end{document}